# PhaseAware: Interpretable Human-in-the-Loop Rehabilitation Scoring with Boundary Monitoring

*Yankai Zheng [1,2#], Yuhe Liu [1,4#], Yuxin Ma [1], Tianci Xue[5], Jiayuan Tian [1], Yu Fu [6], Yuxuan Hu [1,5], Jianing Wang [1], Zichun Xiao[3*], Junya Mu [1*], Shaohui Ma [1*]*

1 Department of Medical Imaging, The First Affiliated Hospital of Xi'an Jiaotong University, Xi'an, 710061, China

2 School of Life Science and Technology, Xi'an Jiaotong University, Xi'an, 710049, China

3 Xi'an North Qinghua Electromechanical Co., Ltd., Xi'an,710025, China

4 Faculty of Electronic and Information Engineering, Xi'an Jiaotong University, Xi'an, 710049, China

5 School of Future Technology, Xi'an Jiaotong University, Xi'an, 710049, China

6 School of Artificial Intelligence, Xidian University, Xi'an, 710126, China

# Yankai Zheng and Yuhe Liu contributed equally to this work.

* Corresponding author:

Shaohui Ma, shh_ma@xjtu.edu.cn

Junya Mu, junyamu@xjtu.edu.cn

Zichun Xiao, 995680173@qq.com

## Abstract

Rehabilitation scoring systems are most useful when their outputs can be reviewed and interpreted within clinical workflows. This study presents PhaseAware, a compact framework for continuous rehabilitation quality assessment that combines a temporal backbone with phase- and body-group descriptors through a backbone-conditioned gated residual pathway. The model was evaluated on the UI-PRMD deep-squat protocol and further tested on the KIMORE squatting subset. On UI-PRMD, PhaseAware achieved an RMSE of 0.0230, corresponding to an 88.9% reduction relative to the accepted baseline. It also maintained favorable performance on KIMORE, suggesting that the phase-aware design transfers across related squatting protocols. In addition to score prediction, PhaseAware generates structured review cues based on phase- and body-level sensitivity, highlighting the movement stages and body regions most relevant to each prediction. The architecture employs a backbone-conditioned gated residual mechanism to stabilize feature representation, supporting use in resource-constrained settings. These cues are intended to support clinician review, boundary-case monitoring, and human-in-the-loop triage rather than autonomous decision-making. Overall, PhaseAware offers a practical and interpretable approach to rehabilitation scoring that may help integrate automated assessment into information systems while preserving clinician oversight.

## Keywords

Action Quality Assessment (AQA); Kinematic Signal Processing; Explainable artificial intelligence; Human-in-the-loop; Remote Rehabilitation Monitoring.

## 1. Introduction

The demand for rehabilitation services continues to increase as musculoskeletal conditions become more prevalent and remote monitoring becomes more widely used [1-4]. In this setting, a movement-quality score has practical value only when it can be stored, queried, audited, and reviewed by clinicians. Rehabilitation assessment is therefore not only a prediction task, but also an information-system task: repeated evaluations must be scalable while remaining open to clinical review.

Action quality assessment (AQA) provides the technical basis for this task. Public datasets such as UI-PRMD and KIMORE have supported reproducible rehabilitation scoring under shared preprocessing and evaluation protocols [7,8], and deep sequence models have been used for continuous quality estimation from kinematic data [5,6,9-15,25]. Prior studies indicate that temporal backbones, attention mechanisms, supervised contrastive learning, sensor-based evaluation, 3D motion tracking, and view-invariant designs can improve benchmark performance. However, many methods still frame rehabilitation assessment as a closed regression task, producing a score without identifying the movement phases or body regions that contributed most to the prediction. For therapists, this limits clinical usefulness because the output offers little guidance clinical triage and objective interpretation, making it difficult to pinpoint anomalous movement phases or prioritize borderline records for expert review.

This limitation is important in rehabilitation because exercises such as the deep squat contain temporally ordered phases and anatomically meaningful body segments that are already used in clinical reasoning. Work on explainable and contestable medical artificial intelligence emphasizes that model outputs should be understandable, actionable, and compatible with human oversight [16-19]. Studies of AI-assisted clinical decision support, modular medical architectures, digital health tools, and workflow visualization also show that clinical systems require components that can be integrated into existing information systems [20-24]. For rehabilitation scoring, a useful model should therefore return not only a bounded quality score, but also concise review cues and boundary-monitoring signals that support human-in-the-loop triage and may conserve clinical bandwidth by directing expert review toward uncertain cases rather than routine high-confidence records [26].This approach transitions from pure black-box regression to a signal-aware, interpretable framework that aligns with contemporary clinical decision support paradigms

We therefore propose PhaseAware, a lightweight rehabilitation scoring component for rehabilitation information systems. We hypothesize that integrating phase- and body-group descriptors through a backbone-conditioned gated residual pathway can improve scoring performance while preserving a clinically readable output interface. PhaseAware combines a temporal backbone with structured phase and body descriptors, allowing the model to retain sequence-learning capacity while producing structured review outputs. It is designed as a human-in-the-loop support tool for score estimation, review prioritization, and boundary monitoring, rather than as an autonomous clinical decision maker. We evaluate PhaseAware on the UI-PRMD deep-squat protocol and further test it on the KIMORE squatting subset to examine predictive performance and cross-dataset generalization. The main contributions are threefold: first, a computationally efficient phase-aware scoring architecture; second, interpretable review outputs correlating quantitative estimates with movement sub-stages; and third, an evaluation that considers both accuracy and information-system integration.

## 2. Methods

### 2.1 Task Definition and Data

The primary task was continuous rehabilitation-quality scoring, first evaluated on the UI-PRMD deep-squat subset [7]. Each sample was represented as a fixed-length kinematic sequence paired with a normalized quality target. Following the preprocessing protocol, each UI-PRMD sequence was linearly resampled to 240 time steps and reorganized into a 90-dimensional grouped representation covering five body segments: trunk, right arm, left arm, right leg, and left leg.

To evaluate cross-protocol generalization while maintaining movement-level comparability, we extended the experiments to Exercise 5 (E5: squatting) in the KIMORE dataset [8]. This subset captures the same basic movement class as the UI-PRMD deep squat but includes greater kinematic variability and a different continuous clinical scoring scale. It therefore provides an external test setting for assessing whether the model remains effective across different rehabilitation scoring protocols.

## 2.2 PhaseAware Model and Review Output

As shown in **Fig. 1**, PhaseAware integrates a convolutional-recurrent temporal backbone with structured auxiliary descriptors to learn sequence-level kinematics and produce clinician-readable review outputs. The aligned motion sequence is divided into three predefined movement windows: descent (frames 0-79), bottom (frames 80-159), and ascent (frames 160-239) [27]. Transition windows were excluded to maintain a compact and reproducible review output.These windows should be interpreted as deterministic pooling regions rather than therapist-annotated biomechanical event boundaries. A visual example of the raw-to-normalized phase standardization used for this fixed-window definition is provided in **Fig. S1**.

Body-group descriptors are aggregated from predefined anatomical channel groups. The phase and body descriptors are projected, normalized, and fused through a backbone-conditioned gated residual pathway, after which the fused representation is passed to a residual-delta scoring head for final score regression. To generate bounded inspection cues, the system applies post-hoc masking within the same inference pipeline, consistent with pose-guided and region-aware feedback strategies in rehabilitation action assessment [28]. This design allows one inference pass to produce a global score and structured phase/body sensitivity maps. The resulting record is intended to support therapist review of boundary cases rather than to provide an autonomous clinical decision.

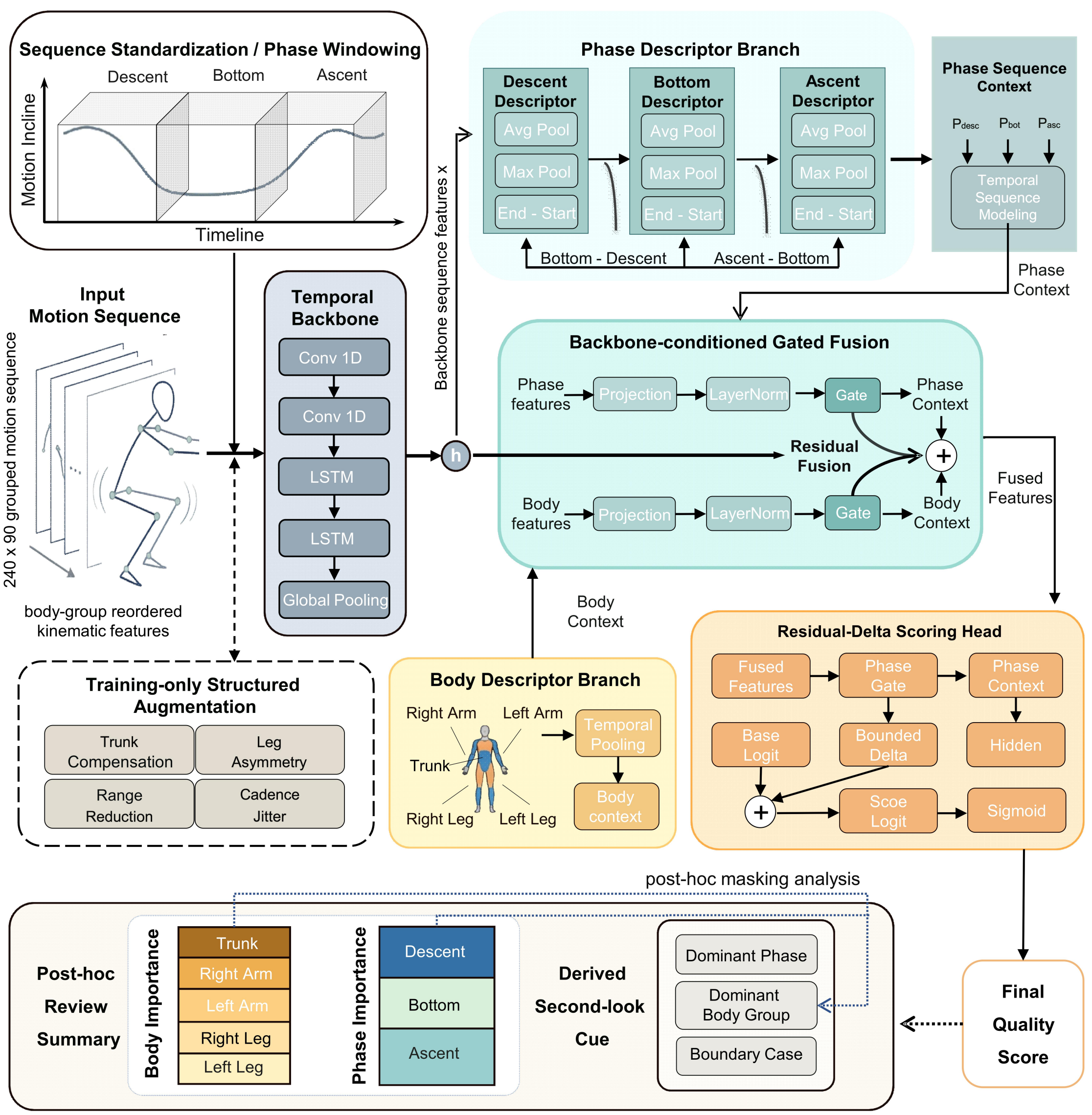

**Fig. 1 Detailed architecture and workflow interface of PhaseAware.** The system processes fixed-length kinematic sequences to regress continuous quality scores and export structured post-hoc review artifacts for human-in-the-loop clinical auditing

## 2.3 Fusion Formulation

Let $X \in \mathbb{R}^{\mathbb{T}\times\mathbb{D}}$ denote the aligned motion sequence, and let $h = \mathsf{GAP}\left(\mathsf{Backbone}(X)\right)$ denote the temporal representation obtained via Global Average Pooling (GAP). To integrate the structural priors, we derive an auxiliary descriptor $k$ for each defined phase window and body-group slice. PhaseAware projects, normalizes, and fuses these descriptors through a backbone-conditioned gated residual pathway. Formally, for each descriptor $k$:

$$\tilde{z}_k = LN(W_k z_k + b_k) \tag{1}$$

$$g_k = \sigma(U_k [h ; \tilde{z}_k] + c_k) \tag{2}$$

$$\tilde{z}_k = g_k \odot \tilde{z}_k \quad (3)$$

where $W_k$ and $U_k$ represent learnable weight matrices, $b_k$ and $c_k$ are bias terms, and $[\cdot\,;\,\cdot]$ denotes channel-wise concatenation. Furthermore, $LN$ represents layer normalization, $\sigma$ is the sigmoid activation function, and $\odot$ denotes the Hadamard (element-wise) product.

The modulated descriptors are subsequently aggregated into a unified auxiliary context $c_{aux}$:

$$c_{aux} = LN(W_{aux}[\hat{z}_1; \dots; \hat{z}_m] + b_{aux}) \quad (4)$$

To preserve the backbone's dominant sequence-learning capacity while incorporating this spatial-temporal prior, the formulation employs a residual-delta design:

$$c_{aux} = \varphi\left(W_{back} + b_{back}\right) \quad (5)$$

$$f = \left[h\,; c_{back} + c_{aux}\right] \quad (6)$$

$$\hat{y} = \sigma(W_o\varphi(W_h f + b_h) + b_o) \quad (7)$$

Here, $\varphi$ denotes the ReLU nonlinearity, and $\hat{y}$ represents the final continuous quality score. This explicit routing ensures that phase and body features act as bounded residual modulators rather than overriding the primary kinematic sequence.

### 2.4 Training, Augmentation, and Feedback Generation

Training used structured movement degradation rather than generic noise injection. Augmented samples were generated from clinically plausible kinematic degradations, including trunk compensation, leg asymmetry, range reduction, and cadence jitter, with bounded severity thresholds and paired label penalties. This augmentation exposed the model to controlled, domain-specific quality variation while preserving the public benchmark comparison protocol.

For the feedback output, the model uses an occlusion-based sensitivity analysis. During inference, each phase window and body-group channel block is masked, and the absolute change in the predicted score is recomputed. The resulting deltas are normalized into a structured contribution summary. These summaries should be interpreted as review heuristics rather than causal explanations: they indicate where a human reviewer may inspect the sequence, but they do not diagnose why a movement is clinically impaired.

## 3. Experimental Setup

All main experiments used the same optimization protocol. The dataset was partitioned into a 70/30 train-evaluation split. Models were optimized with Adam using an initial learning rate of 1e-4, a batch size of 16, and a sequence dropout rate of 0.25. Training was run for 12 epochs with a binary cross-entropy objective function.

To account for initialization variance, all models were trained and evaluated across five predefined random seeds (1337, 3141, 5151, 1919, 2727). For comparison, we retained two reference anchors. The accepted baseline represents historical benchmark performance, whereas the clean comparator provides a conservative parameter-matched reference. These anchors are used for interpretive comparison and should not be treated as matched rerun models. The reference anchors are summarized in **Table 1**.

**Table 1 Performance comparison against historical and structural reference anchors**

| Reference | RMSE | MAD |
|---|---|---|
| Accepted baseline anchor | 0.2074 | 0.1979 |
| Clean matched-budget comparator | 0.0536 | 0.0429 |
| PhaseAware | 0.0230 ± 0.0018 | 0.0159 ± 0.0006 |

Data for PhaseAware are presented as mean ± SD computed over the five predefined initialization seeds

### 3.1 Evaluation Metrics

The primary endpoint was the root mean squared error (RMSE) evaluated on the held-out evaluation set. Mean absolute deviation (MAD), the coefficient of determination ($R^2$), and Bland-Altman limits of agreement (LoA) were incorporated as complementary metrics to capture distinct dimensions of predictive performance: average magnitude of error, robust absolute deviation, explained variance, and clinical agreement-band width, respectively.

The metrics are formally defined as follows:

$$RMSE = \sqrt{\frac{1}{n}\sum_{i=1}^{n}\left(\widehat{y_i} - y_i\right)^2} \quad (8)$$

$$\mathrm{MAD} = \frac{1}{n}\sum_{i=1}^{n}\left|\widehat{y_i} - y_i\right| \quad (9)$$

$$R^2 = 1 - \frac{\sum_{i=1}^{n}\left(y_i - \widehat{y_i}\right)^2}{\sum_{i=1}^{n}\left(y_i - \overline{y}\right)^2} \quad (10)$$

$$LoA = \overline{d} \pm 1.96 s_d, \quad d_i = \widehat{y_i} - y_i \quad (11)$$

Here, $y_i$ represents the reference clinical quality score, $\widehat{y_i}$ denotes the model-predicted score, and $n$ is the total number of evaluated samples. For the variance and agreement analyses, $\overline{y}$ denotes the mean of the reference scores, $d_i$ defines the individual prediction difference, $\overline{d}$ is the mean prediction bias, and $s_d$ signifies the standard deviation of these differences.

### 3.2 Governance and Intended Use Boundary

This study used previously released, de-identified public datasets and did not involve new human-subject recruitment or collection of identifiable patient data. The research scope is therefore systems-oriented and preclinical. PhaseAware is evaluated as a computational scoring and review-prioritization component, not as an autonomous clinical assessor. For future clinical integration, the minimum structured record should include three elements: the normalized performance score, review cues derived from dominant phase and body-group sensitivities, and prioritization flags such as score-bin assignment, boundary-monitoring status, and suggested review targets. This record-oriented format is designed to support storage, query, audit, and longitudinal tracking without requiring clinicians to interpret latent neural representations directly.

## 4. Results

### 4.1 Main Quantitative and Ablation Results

As summarized in **Table 2** and visualized in **Fig. 2**, PhaseAware consistently achieved superior predictive accuracy to all established benchmark models. On the held-out UI-PRMD deep-squat protocol, PhaseAware achieved lower prediction error than the reference models, with a mean RMSE of 0.0230 ± 0.0018 and an $R^2$ of 0.8442 ± 0.0267. Notably, the tight error intervals—indicated by the standard deviation (SD) across five random seeds in Fig. 2— underscore the model's stable convergence.  Metric-specific supplementary comparisons for RMSE, MAD, $R^2$, and LoA width are provided in **Figs. S2-S5**. Seed-level metric profiles for PhaseAware and the comparator models are shown in Figs. **S6-S12**.

**Table 2 Quantitative performance comparison.**

| Method | RMSE | MAD | R² | LoA width |
|---|---|---|---|---|
| PhaseAware | 0.0230 ± 0.0018 | 0.0159 ± 0.0006 | 0.8442 ± 0.0267 | 0.0878 ± 0.0057 |
| InceptionTime | 0.0326 ± 0.0052 | 0.0237 ± 0.0049 | 0.6773 ± 0.1277 | 0.1090 ± 0.0148 |
| ResNet1D | 0.0350 ± 0.0075 | 0.0275 ± 0.0062 | 0.6431 ± 0.1070 | 0.1094 ± 0.0156 |
| TCN | 0.0400 ± 0.0073 | 0.0296 ± 0.0061 | 0.5367 ± 0.1013 | 0.1529 ± 0.0285 |
| ST-Transformer | 0.0422 ± 0.0065 | 0.0323 ± 0.0048 | 0.4806 ± 0.1074 | 0.1157 ± 0.0178 |
| SpatioTemporalNN | 0.0460 ± 0.0109 | 0.0362 ± 0.0081 | 0.3727 ± 0.2192 | 0.1793 ± 0.0438 |
| MLP | 0.0472 ± 0.0082 | 0.0334 ± 0.0060 | 0.3475 ± 0.1544 | 0.1676 ± 0.0267 |
| CNN | 0.0522 ± 0.0081 | 0.0393 ± 0.0051 | 0.2022 ± 0.1685 | 0.1974 ± 0.0347 |

Values for PhaseAware are presented as mean ± SD across five independent initialization seeds. Baselines are included as reference anchors for historical performance and architecture-level comparison

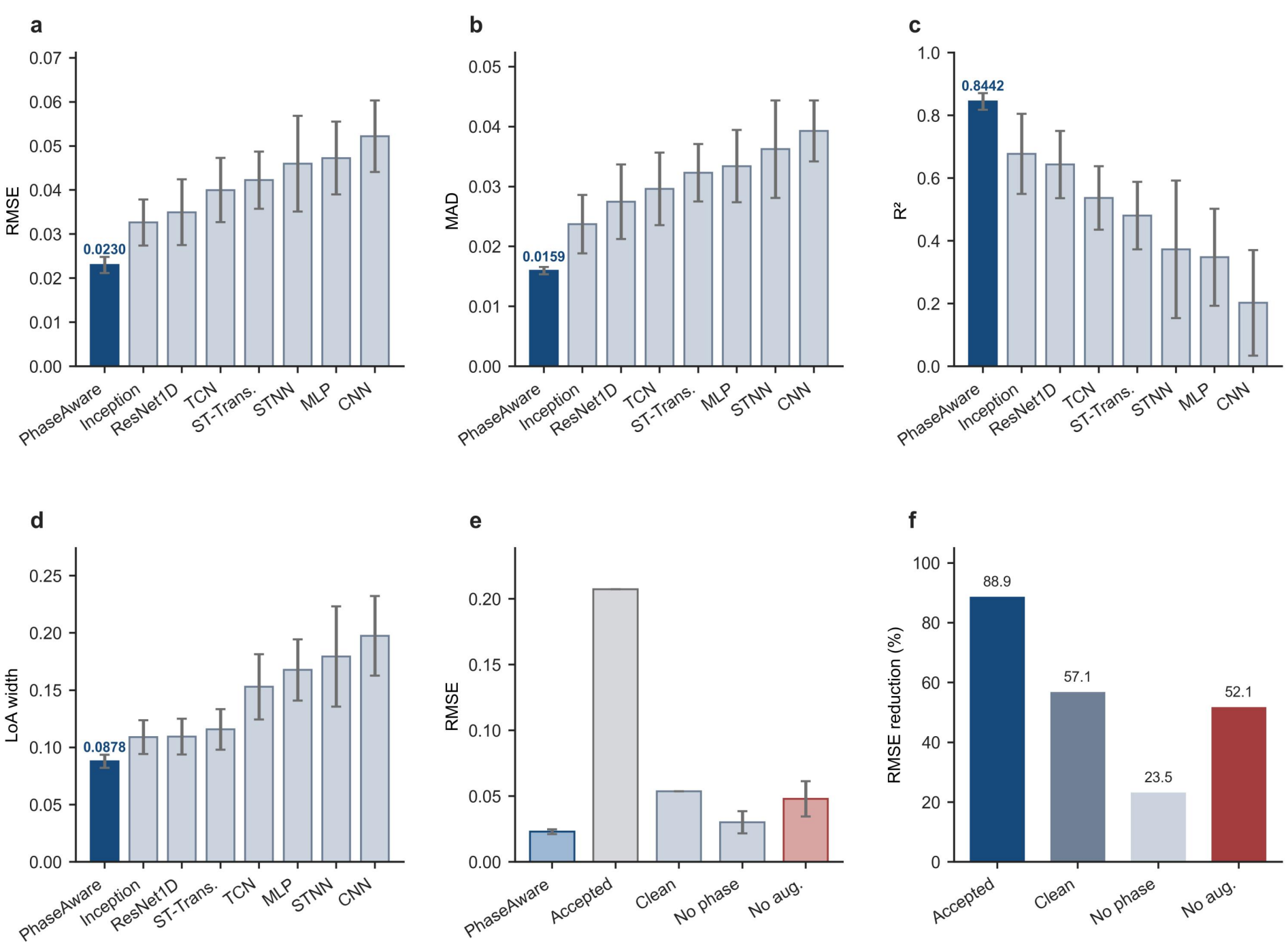

Fig. 2 Predictive performance and ablation comparison across six panels. a RMSE, b MAD, c R², and d LoA width across PhaseAware and comparator models. e Absolute RMSE comparison against the accepted anchor, clean comparator, no-phase ablation, and no-augmentation ablation. f Relative RMSE reduction against the same comparison anchors. Error bars represent SD across five independent initialization seeds when available.

The combined Fig. 2e-f further summarizes relative error reductions and ablation results. Compared with the accepted baseline anchor, PhaseAware reduced RMSE by 88.9%; compared with the clean matched-budget comparator, it reduced RMSE by 57.1%. In the ablation analysis, removing the phase branch increased RMSE to 0.0301 ± 0.0084, corresponding to a 23.5% relative performance reduction under the predefined seed package, while removing structured augmentation increased RMSE to 0.0480 ± 0.0134. Additional ablation sensitivity plots and seed-level mechanism deltas are reported in Figs. S13-S16.

### 4.2 Generalization Performance on KIMORE Dataset

To examine whether PhaseAware generalizes beyond the UI-PRMD protocol, we evaluated it on the KIMORE dataset. Testing was limited to Exercise 5 (E5: squatting) to maintain movement-level comparability while introducing greater kinematic variability. As shown in **Table 3**, PhaseAware achieved the lowest mean RMSE (0.0973 ± 0.0115) and the highest $R^2$ (0.7245 ± 0.0464) among the evaluated models. Because KIMORE uses a different continuous clinical scoring scale, the absolute error values are not directly comparable with those from UI-PRMD; however, the relative ranking suggests that the phase-aware architecture retained favorable performance across protocols.

**Table 3 Cross-protocol generalization performance on the KIMORE (Exercise 5: squatting) dataset.**

| Model | RMSE | MAD | $R^2$ | LoA width |
|---|---|---|---|---|
| PhaseAware | 0.0973 ± 0.0115 | 0.0747 ± 0.0090 | 0.7245 ± 0.0464 | 0.3718 ± 0.0543 |
| MLP | 0.0987 ± 0.0113 | 0.0773 ± 0.0081 | 0.7131 ± 0.0746 | 0.3736 ± 0.0299 |
| ST-Transformer | 0.1085 ± 0.0107 | 0.0815 ± 0.0100 | 0.6546 ± 0.0754 | 0.4102 ± 0.0402 |
| SpatioTemporalNN | 0.1273 ± 0.0118 | 0.0997 ± 0.0097 | 0.5295 ± 0.0595 | 0.4908 ± 0.0440 |
| TCN | 0.1288 ± 0.0163 | 0.0997 ± 0.0100 | 0.5042 ± 0.1576 | 0.4862 ± 0.0519 |
| InceptionTime | 0.1312 ± 0.0070 | 0.0995 ± 0.0098 | 0.4987 ± 0.0596 | 0.5070 ± 0.0259 |
| ResNet1D | 0.1449 ± 0.0195 | 0.1138 ± 0.0160 | 0.3734 ± 0.1938 | 0.5089 ± 0.0483 |
| CNN | 0.1494 ± 0.0157 | 0.1242 ± 0.0174 | 0.3506 ± 0.1042 | 0.5660 ± 0.0559 |

Values are computed across five independent initialization seeds and presented as mean ± SD

### 4.3 Agreement and Error Structure

The agreement and error structure in Fig. 3 uses ensemble-averaged predictions across the five evaluation seeds. In the prediction-reference plot (Fig. 3a), high-, middle-, and low-performance strata are distributed along the identity line. Bland-Altman analysis (Fig. 3b) showed a mean bias of 0.0023, with limits of agreement from -0.0416 to 0.0462. The agreement interval covered 94.3% ± 1.5% of observations. The absolute-error distributions (Fig. 3c) and score-band error bars (Fig. 3d) indicate heteroscedastic error behavior, with lower variance for high-quality movements and wider variance in the low- and middle-score strata.

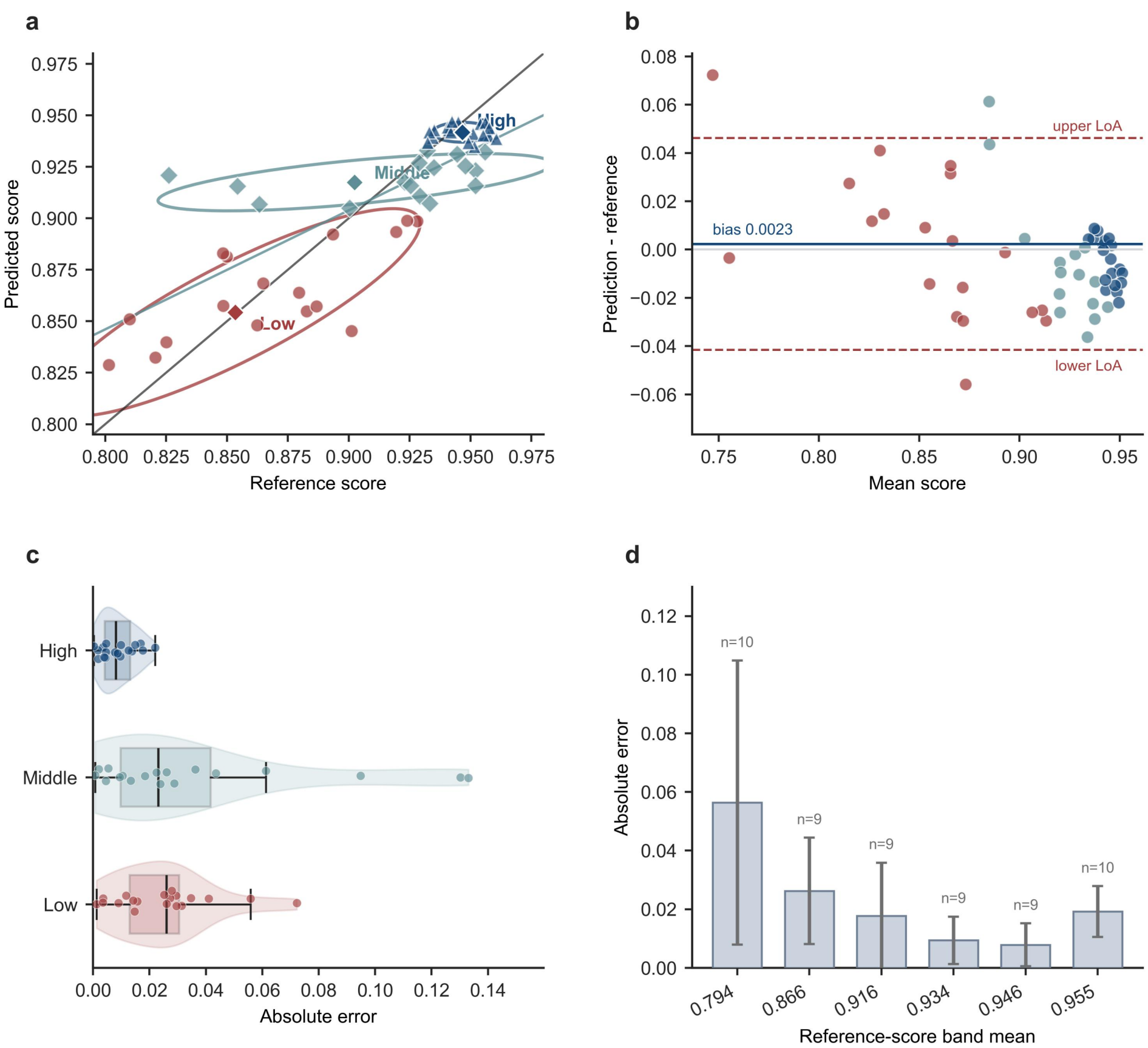

Fig. 3 Consistency and error distribution across clinical score strata. Predicted values for all panels represent the ensemble average across five evaluation seeds. a Prediction-reference geometry highlighting three score-bin clusters with covariance ellipses. b Bland-Altman analysis of the ensemble predictions. c Absolute-error distributions stratified by score bin. d Score-band mean absolute error summarized with vertical bars and SD error bars across reference-score bands.

## 4.4 Review Utility and Boundary Behavior

Beyond prediction accuracy, a scoring component intended for clinical use should provide review cues that clinicians can inspect. As shown in Fig. 4a, PhaseAware preserved ordinal consistency across score strata: mean reference scores aligned with predictions, increasing from 0.8401 to 0.9481, while mean absolute error decreased. The pairwise ranking accuracy was 0.935 ± 0.014. These findings support the use of the output as a triage and review-prioritization signal, not as an autonomous clinical endpoint.

The model's boundary behavior showed patterns that were consistent with clinical expectations. For high-scoring sequences, the model heavily concentrates its body-group weighting on the lower extremities (0.895), appropriately focusing on the primary biomechanical drivers of a deep squat (Fig. 4d). Conversely, for low-scoring movements, this attention redistributes, exhibiting a significant 0.339 weighting shift toward the upper body (trunk and arms). This shift suggests that the network successfully learns to identify the upper-body compensation strategies typical of degraded, pathological squat mechanics.

The highest-error cases also provide a basis for defining human-in-the-loop review priorities. Prediction errors were concentrated in the ascent and bottom phases (Fig. 4b and Fig. 4e) and were accompanied by leg-dominant body cues (Fig. 4c and Fig. 4f). These patterns can help identify movement windows that may require therapist inspection when the model output is uncertain. The expanded main figure incorporates the phase- and body-vote heatmaps that were previously presented separately in the appendix; the standalone body-weight supplementary heatmap is retained in Fig. S17.

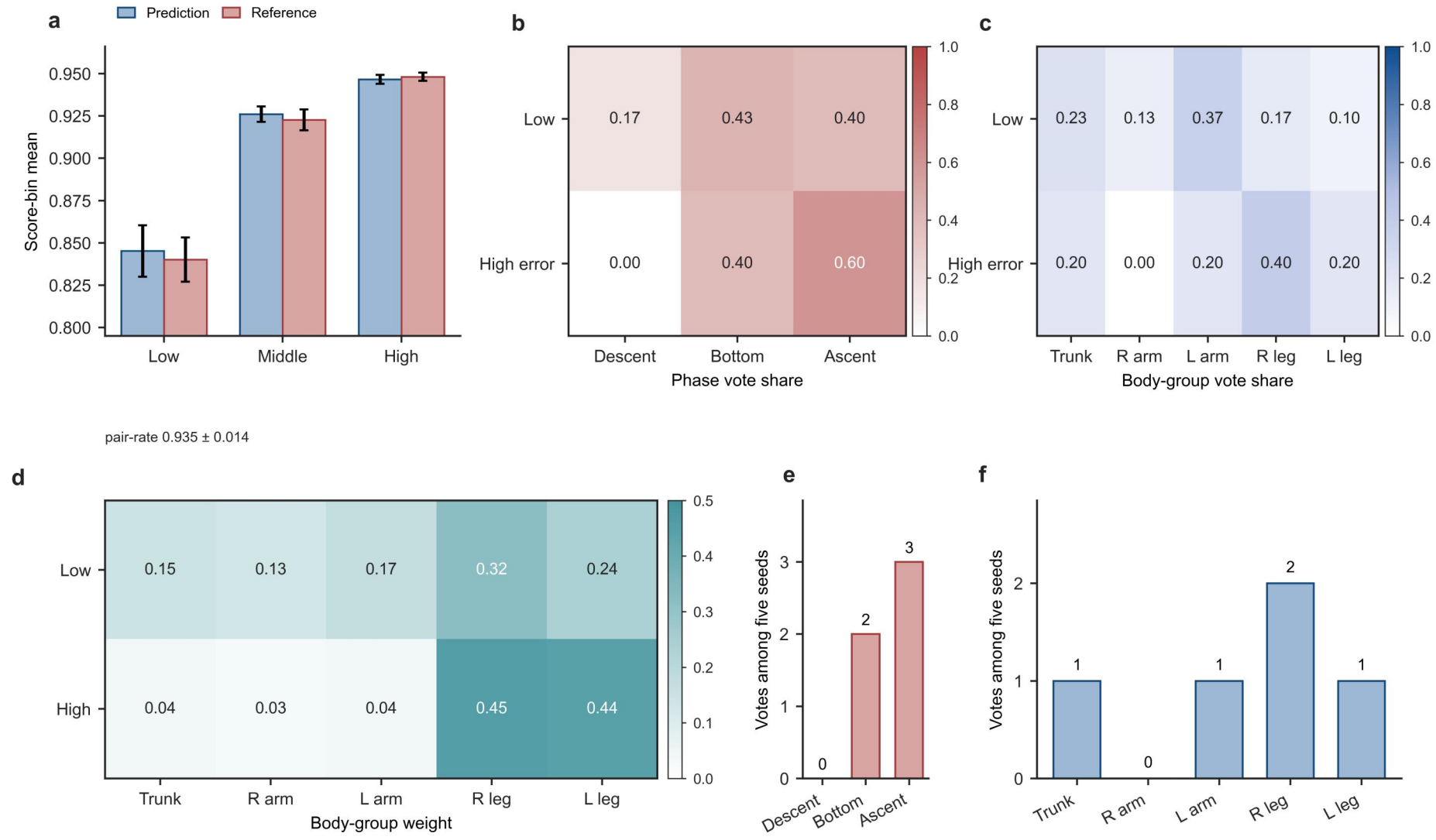

Fig. 4 Clinical review utility and boundary monitoring artifacts. All represented data reflect outputs across the five predefined evaluation seeds. a Ordinal consistency of prediction and reference score-bin means shown as vertical bars with SD error bars. b Phase-window vote distribution for low-score and highest-error cases. c Body-group vote distribution for low-score and highest-error cases. d Body-group weighting shifts between low- and high-scoring movements. e Highest-error phase vote concentration. f Highest-error body-group vote concentration.

### 4.5 System Profile and Integration Path

The proposed PhaseAware component was implemented using the PyTorch framework and profiled locally on a consumer-grade, ARM-based mobile processor (Apple M2 Silicon) to validate its suitability for resource-constrained clinical environments. Under these standard CPU constraints, the model exhibits a compact computational footprint of 170,721 trainable parameters and a total size of 2.4 MB. The inference architecture supports asynchronous batch processing, with a measured latency of 7.26 ms per kinematic sample and a total throughput of 54.9 sequences per second for fixed-length skeleton inputs (450 frames).

### 4.6 Limitations and Future Work

While the proposed architecture demonstrates robust performance, its clinical utility is currently validated primarily on the UI-PRMD and KIMORE datasets. Future work will focus on cross-center validation to assess the generalizability of the gated residual mechanism across more heterogeneous clinical environments and long-term rehabilitation monitoring workflows.

## 5. Discussion

### 5.1 Principal Findings

This study developed PhaseAware as a lightweight rehabilitation scoring component that links continuous score estimation with structured review cues and boundary monitoring. The main finding is that a backbone-conditioned residual phase branch yielded superior estimation accuracy while maintaining a clinically-aligned, interpretable interface suitable for integration. Compared with conventional sequence models and structural baselines, PhaseAware produced lower prediction error on the UI-PRMD deep-squat protocol and retained favorable performance on the KIMORE squatting subset. The model output was not limited to a scalar score; it also provided

phase- and body-group sensitivity patterns that may help therapists identify movement regions requiring closer inspection.These performance gains indicate that explicit modeling of phase-level dependencies effectively mitigates the non-linear noise inherent in kinematic time-series data.

## 5.2 Methodological and Clinical Interpretation

These findings are relevant because rehabilitation quality assessment is not a generic time series regression problem. Rehabilitation movements include ordered phases and anatomically meaningful body segments that are already part of clinical reasoning. Previous studies have shown that deep models can estimate rehabilitation scores from kinematic data [9-15], but many approaches focus mainly on benchmark accuracy. PhaseAware extends this work by embedding predefined phase windows and body-group descriptors into the scoring pipeline. In this design, the temporal backbone learns the main sequence representation, whereas the structured branch contributes bounded auxiliary information. This reduces the risk that manually defined descriptors dominate learned kinematic features and makes the output easier to review in a therapist-facing setting.

The agreement and error analyses clarify the intended interpretation of the model. The near-zero mean bias and narrow limits of agreement indicate that PhaseAware does not show a strong systematic tendency to overestimate or underestimate quality scores. However, the wider error variance in low- and middle-score strata suggests that impaired or non-standard movements remain more difficult to model than high-quality executions. This pattern is clinically plausible, because pathological or compensatory movement can vary across subjects even when the nominal exercise label is identical. Therefore, PhaseAware should not be interpreted as an autonomous assessor. Its more appropriate use is review prioritization: high-confidence cases may support efficient screening, whereas boundary cases should be routed to human inspection.

## 5.3 Implications for Explainable Rehabilitation AI Systems

The review cues generated by PhaseAware should be interpreted as inspection cues rather than causal explanations. This distinction is important for clinical AI governance. Explainable AI literature emphasizes that medical AI outputs should be understandable, contestable, and linked to user action [16-19]. In this context, PhaseAware does not identify the biomechanical cause of a low score; instead, it narrows the reviewer's attention to specific temporal and anatomical regions, consistent with recent work on pose-guided rehabilitation assessment and dual-referenced action-quality feedback [28,29]. This type of output is better aligned with clinical decision support than a score alone, particularly in remote rehabilitation systems where therapists may need to triage repeated exercise records.

The model's small parameter count and low inference latency also support potential integration into rehabilitation information systems. Recent studies on AI-assisted clinical decision support and modular medical architectures highlight the need for components that can be stored, queried, audited, and incorporated into existing workflows [20,22]. PhaseAware addresses this requirement by generating a structured record that includes a normalized score, dominant review cues, and boundary-monitoring indicators. This record-oriented output may support longitudinal tracking and retrospective audit without requiring clinicians to interpret latent neural representations directly.

## 5.4 Limitations and Future Work

Several limitations should be considered before clinical translation. First, the current evidence is based on public datasets and controlled squatting protocols, including UI-PRMD and KIMORE [7,8]. Although cross-dataset evaluation supports sequence-level generalization, broader validation is needed across additional exercises, sensor settings, patient groups, and rehabilitation stages. Second, the phase windows were defined as deterministic pooling regions rather than therapist-annotated biomechanical events. This improves reproducibility but may not capture subject-specific timing variation. Future work should compare fixed windows with clinically annotated or adaptively detected movement phases. Third, the review utility of the exported cues has not yet been tested in prospective therapist-in-the-loop studies. A necessary next step is to determine whether these cues reduce audit time, improve reviewer confidence, or improve clinical decision consistency.

Overall, PhaseAware provides a compact and auditable framework for AI-assisted rehabilitation scoring. Its main scientific value lies in combining score accuracy, interpretable review output, and boundary-aware workflow design within a single model interface. The current results support PhaseAware as a preclinical research component for rehabilitation information systems and identify the validation steps needed before use in real clinical environments.

## 6. Conclusion

This study aimed to develop and evaluate PhaseAware, a lightweight rehabilitation scoring framework designed to integrate continuous movement-quality estimation with interpretable review artifacts and boundary monitoring. We hypothesized that embedding phase- and body-group information through a backbone-conditioned residual pathway would improve scoring performance while preserving a clinically readable output interface.

The results support this hypothesis. PhaseAware achieved strong performance on the UI-PRMD deep-squat protocol and maintained favorable performance on the KIMORE squatting subset, suggesting that structured phase-aware modeling can improve rehabilitation quality assessment beyond conventional sequence-based approaches. In addition to score prediction, the model generated phase- and body-level review cues, allowing the output to serve as a review-prioritization signal rather than a standalone clinical judgment. This design supports the development of rehabilitation AI systems that are accurate, auditable, workflow-compatible, and aligned with human-in-the-loop clinical review.

Several limitations remain. The current evaluation was based on public datasets and controlled squatting protocols, and prospective validation in broader rehabilitation exercises, diverse patient populations, and real clinical workflows is still required. The fixed phase windows also provide reproducibility but may not fully capture individualized movement timing. Future studies should examine adaptive or clinician-annotated phase definitions and assess whether the exported review artifacts can reduce therapist audit burden and improve decision consistency.

Overall, PhaseAware provides a compact and interpretable framework for AI-assisted rehabilitation scoring and may support clinically supervised rehabilitation information systems.

## Statements and Declarations

### Funding

The study was supported by the Health Research and Innovation Capacity Strengthening Platform Program of Shaanxi Province (Grant Number 2023PT-09)

### Declaration of competing interest

The authors have no relevant financial or non-financial interests to disclose.

### Ethics Approval

This study used previously published public datasets and did not involve new human-subject recruitment, intervention, or identifiable patient data collection by the authors.

### Consent to Participate

Not applicable.

### Consent to Publish

Not applicable.

### Data Availability

The primary datasets are public UI-PRMD and KIMORE. Processed experiment outputs are derived from those sources under the train/validation protocols described in Methods.

### Code Availability

Training configuration and analysis scripts are available from the corresponding author on reasonable request, subject to dataset license constraints.

### Author Contributions

Y.Z.: Conceptualization, Methodology, Software, Formal analysis, Visualization, Writing – Original Draft. Y.L.: Data curation, Validation, Writing – Review & Editing. Y.M., T.X., J.T., Y.F., Y.H., J.W.: Data curation, Investigation. Z.X.: Resources, Writing – Review & Editing. J.M.: Supervision, Project administration, Writing – Review & Editing. S.M.: Supervision, Project administration, Funding acquisition, Writing – Review & Editing.

### AI Use Disclosure

Generative AI tools were used for language restructuring, formatting assistance, and figure-generation scripting. No scientific claim, metric, reference, or conclusion was accepted without verification against local experiment records.

Clinical trial number: not applicable.

# Supplementary Information

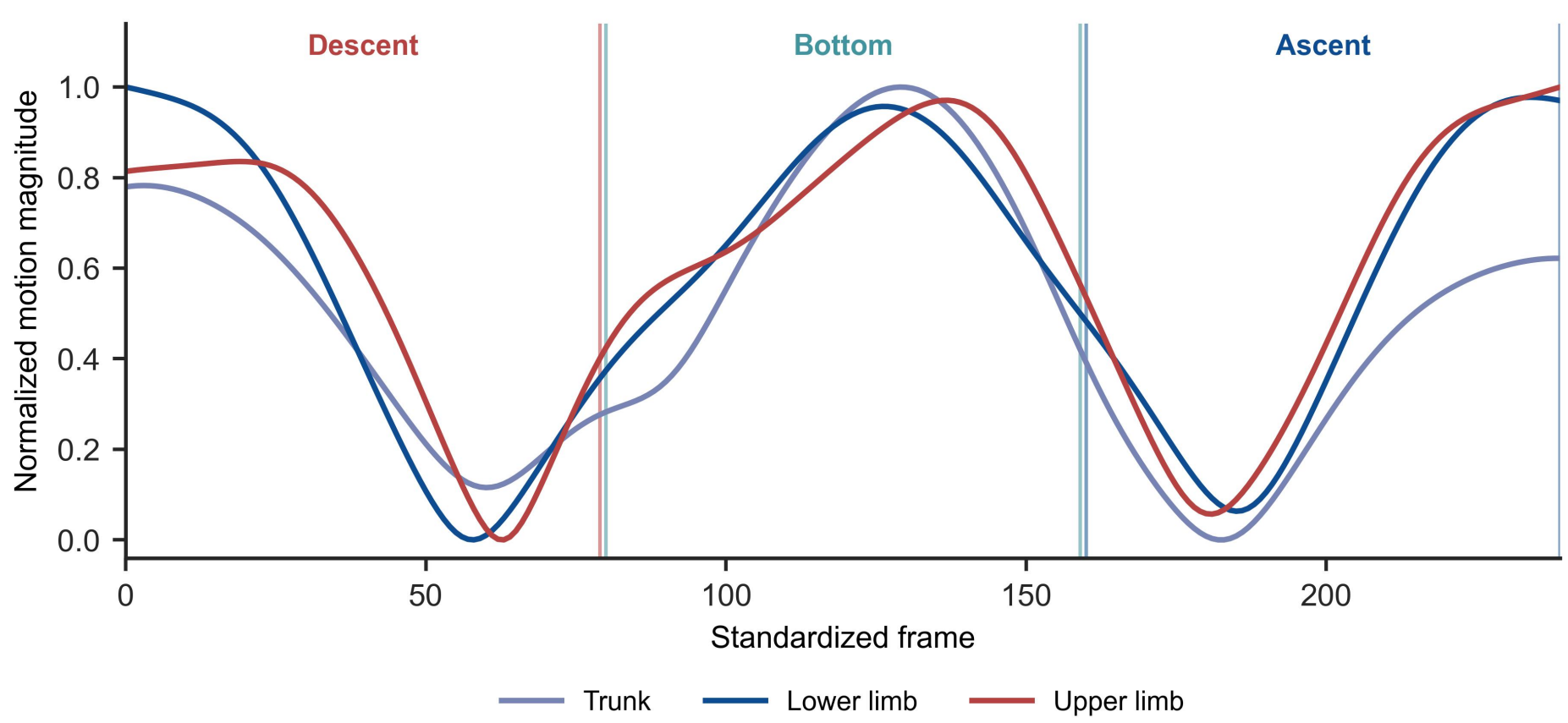

**Fig. S1** Phase standardization example. Raw and normalized views are retained as a visual reference for the fixed-window phase definition used in the main manuscript

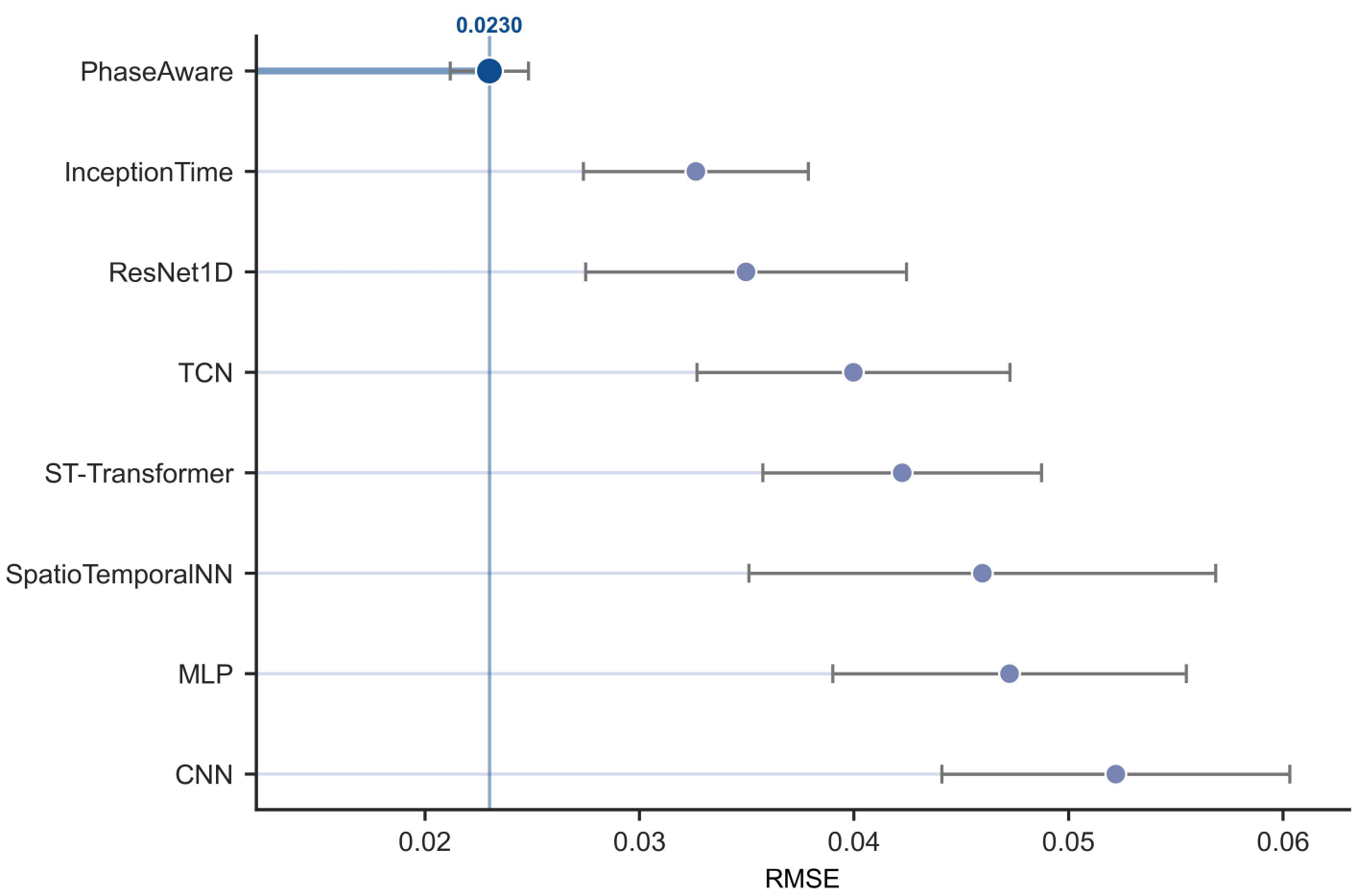

**Fig. S2** Supplementary RMSE model comparison. Mean ± SD over the five predefined seeds; lower values are better

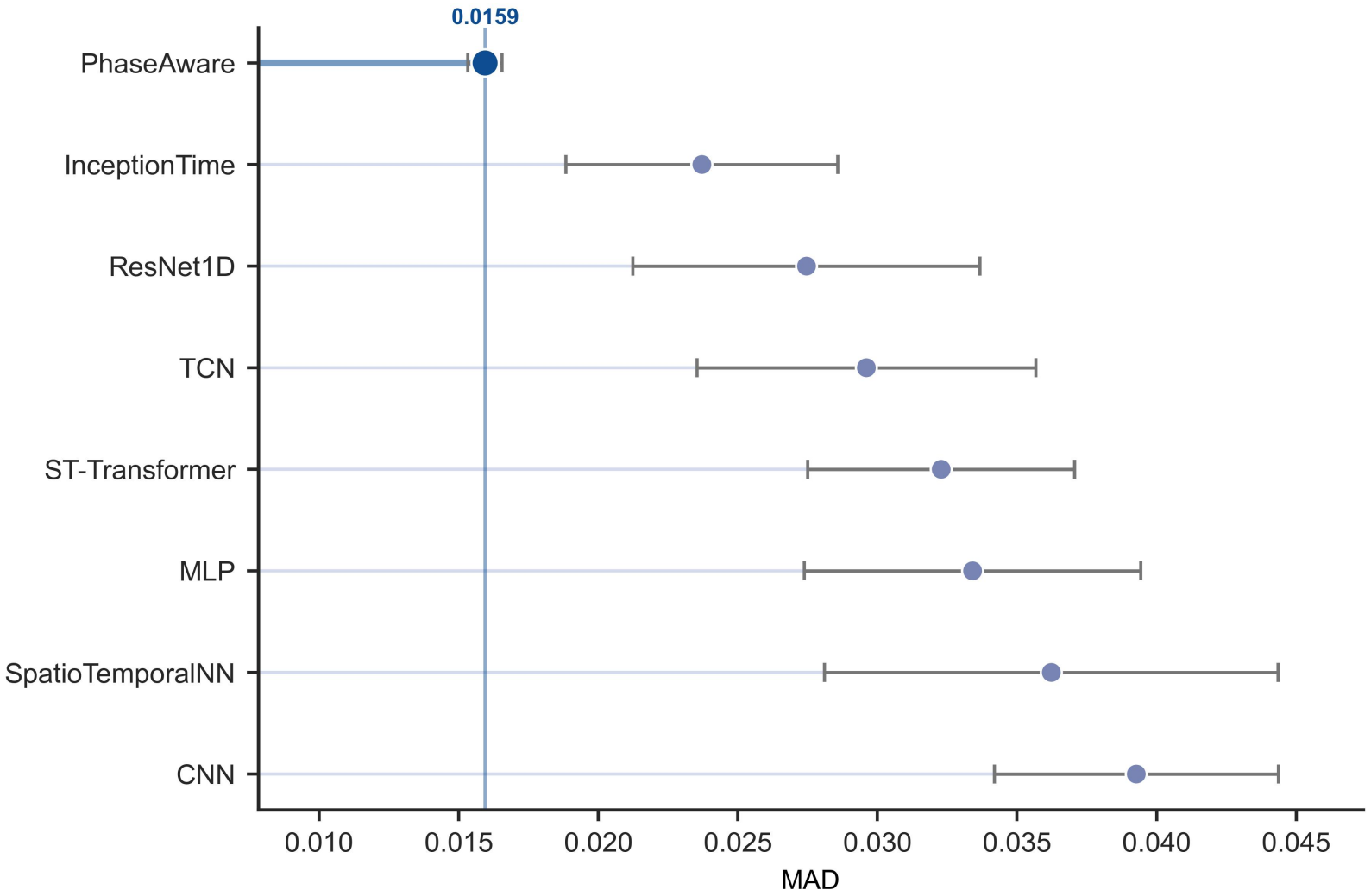

**Fig. S3** Supplementary MAD model comparison. Mean ± SD over the five predefined seeds; lower values are better

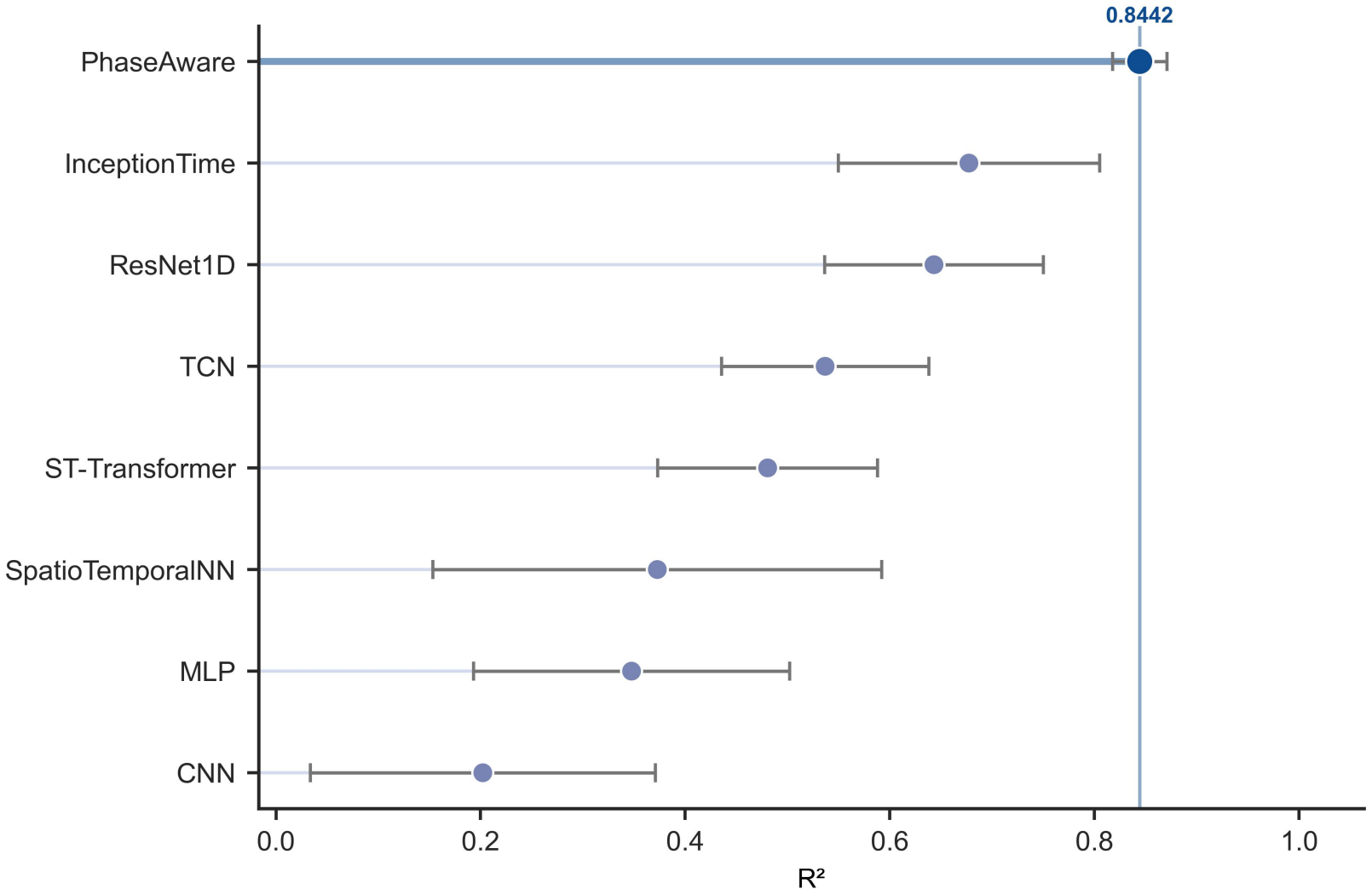

**Fig. S4** Supplementary $R^2$ model comparison. Mean ± SD over the five predefined seeds; higher values are better

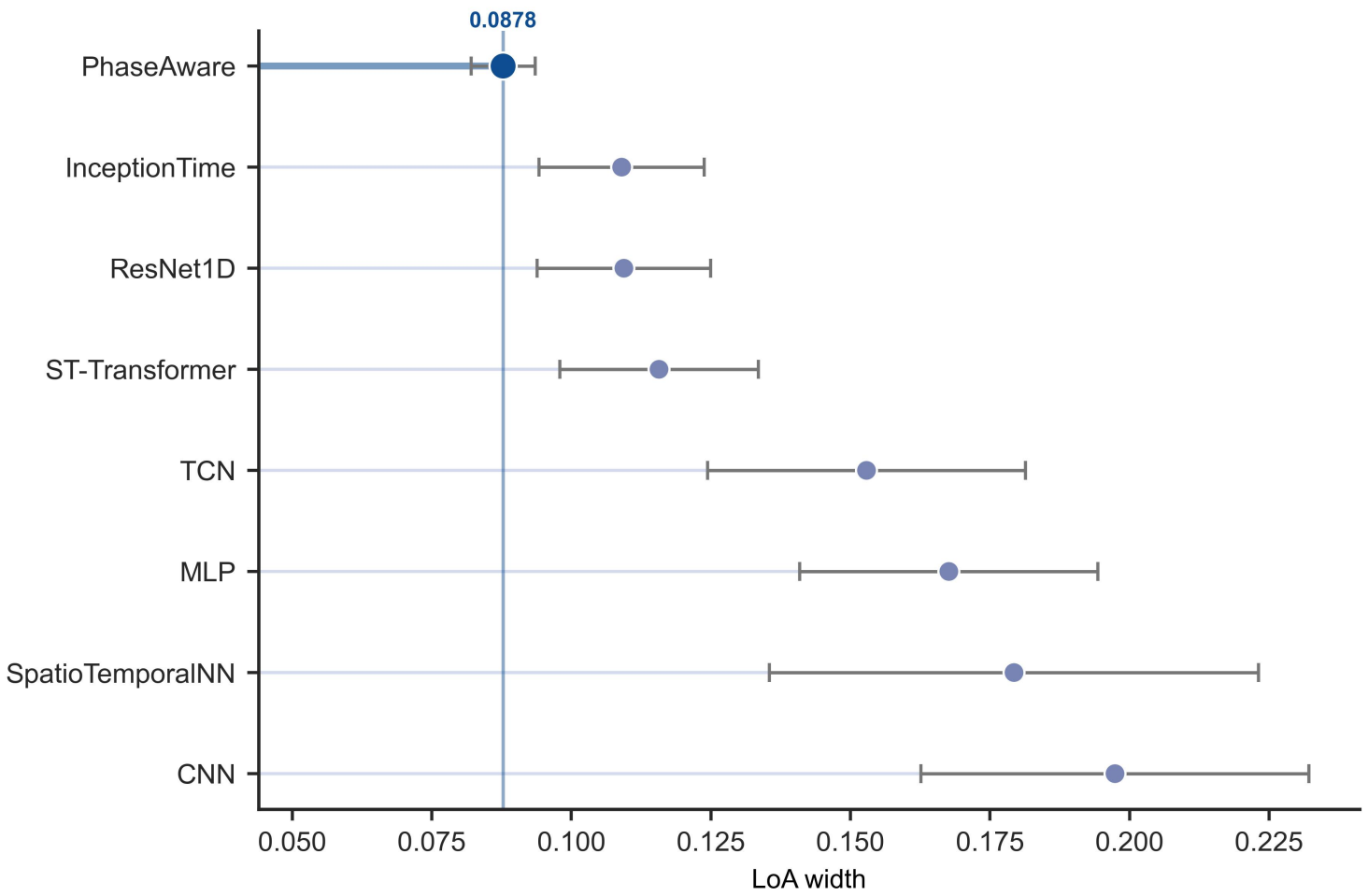

**Fig. S5** Supplementary LoA-width model comparison. Mean ± SD over the five predefined seeds; lower values are better

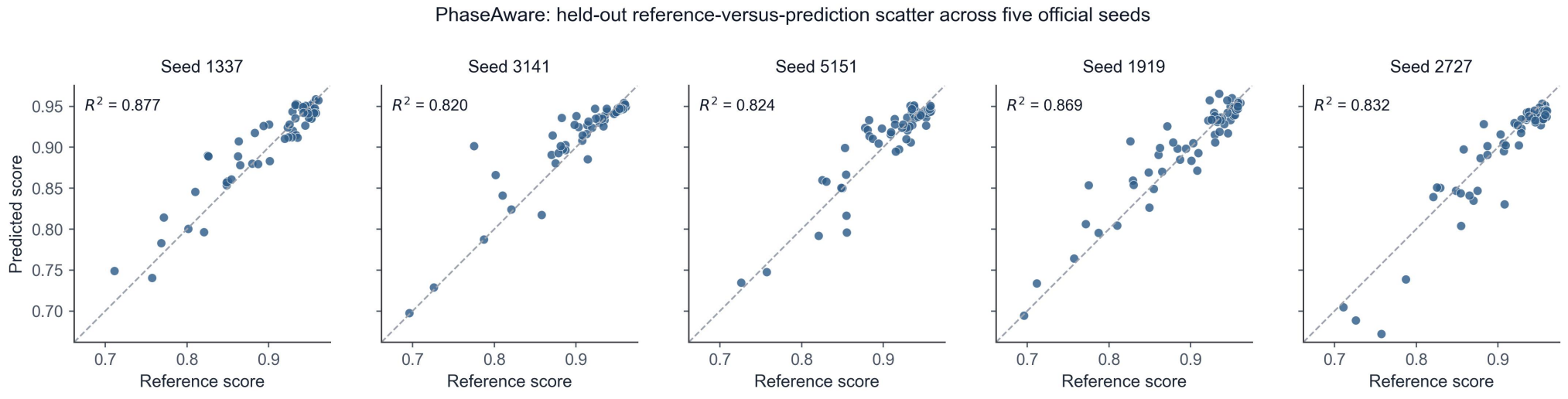

**Fig. S6 PhaseAware held-out reference-versus-prediction scatter across the five official seeds, with per-panel $R^2$ annotations.**

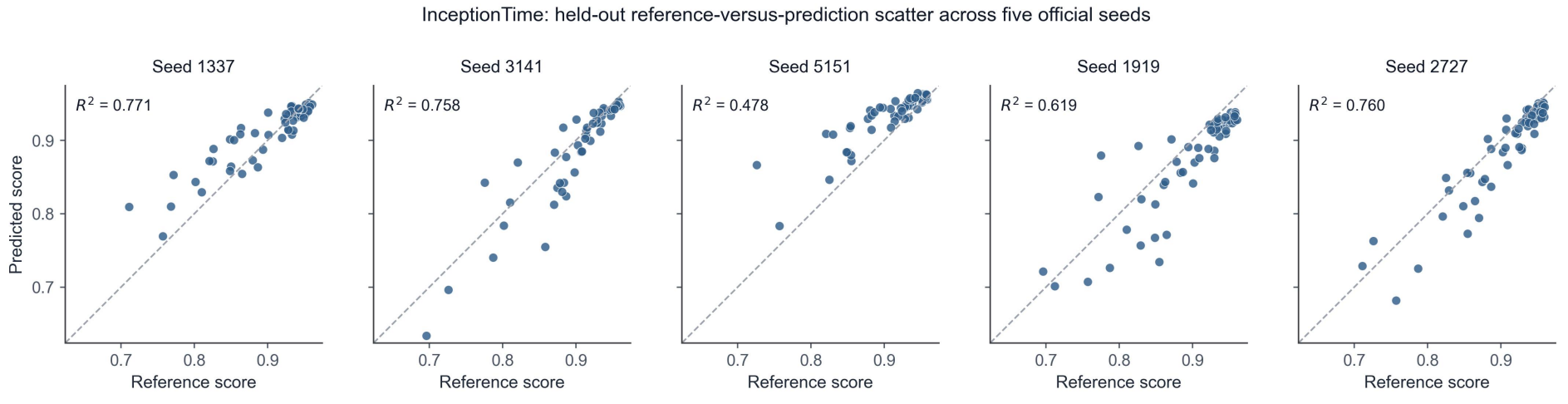

**Fig. S7 InceptionTime held-out reference-versus-prediction scatter across the five official seeds, with per-panel $R^2$ annotations.**

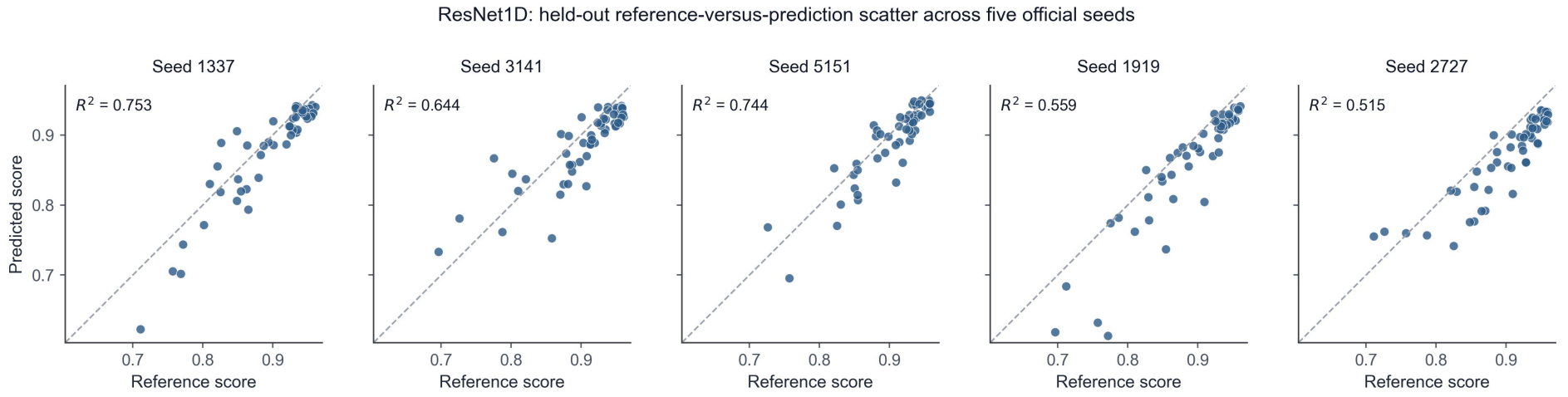

**Fig. S8 ResNet1D held-out reference-versus-prediction scatter across the five official seeds, with per-panel $R^2$ annotations.**

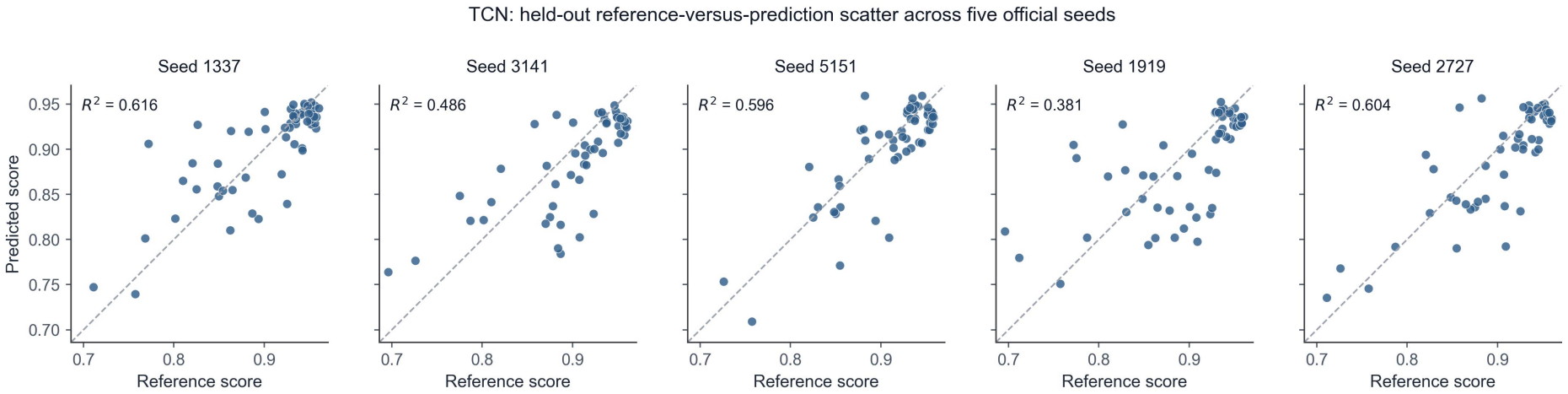

**Fig. S9 TCN held-out reference-versus-prediction scatter across the five official seeds, with per-panel $R^2$ annotations.**

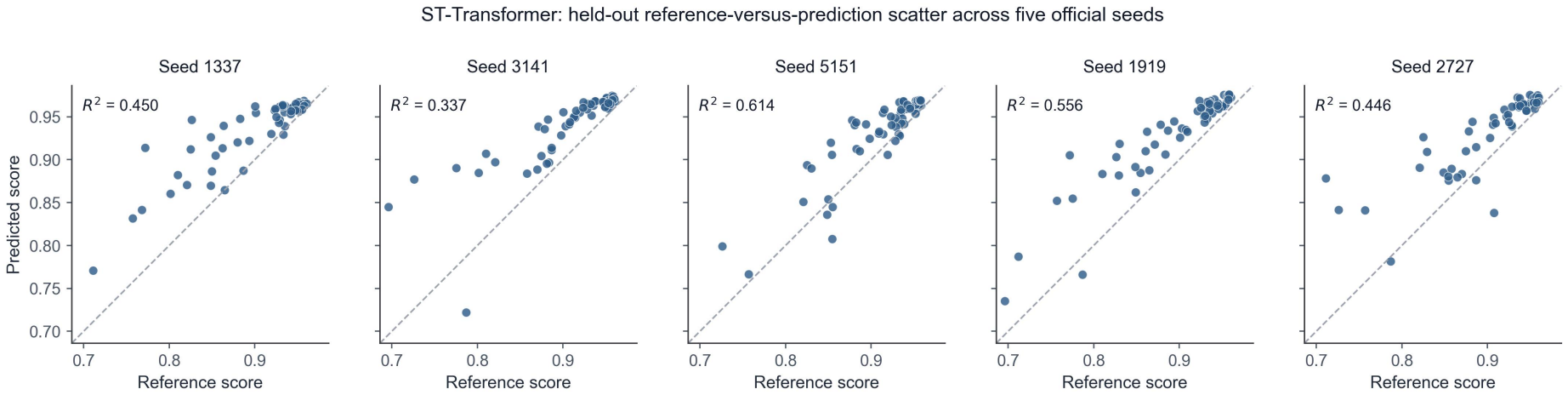

**Fig. S10 ST-Transformer held-out reference-versus-prediction scatter across the five official seeds, with per-panel $R^2$ annotations.**

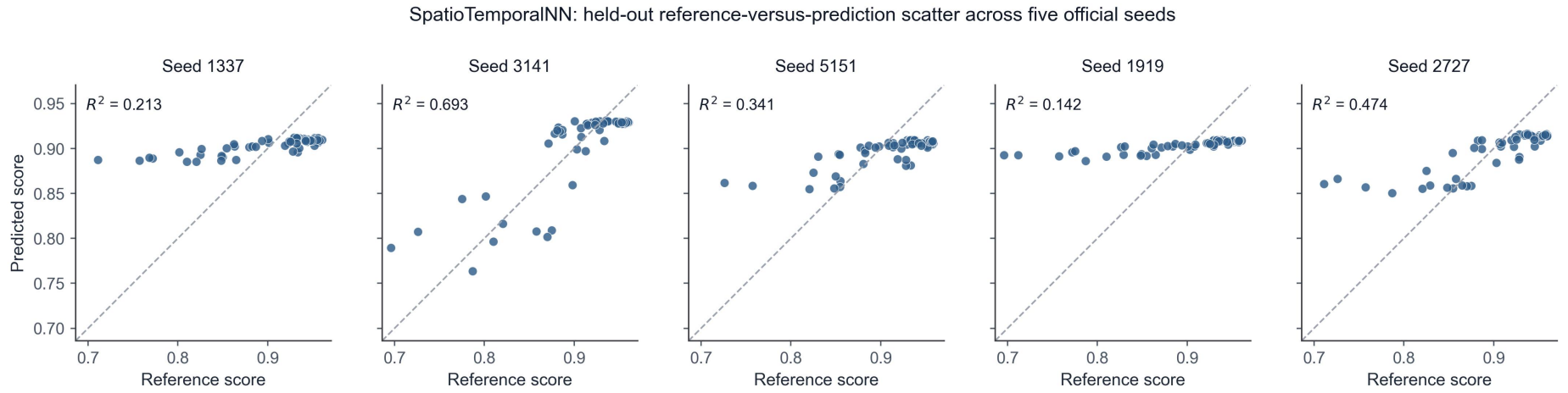

**Fig. S11 SpatioTemporalNN held-out reference-versus-prediction scatter across the five official seeds, with per-panel $R^2$ annotations.**

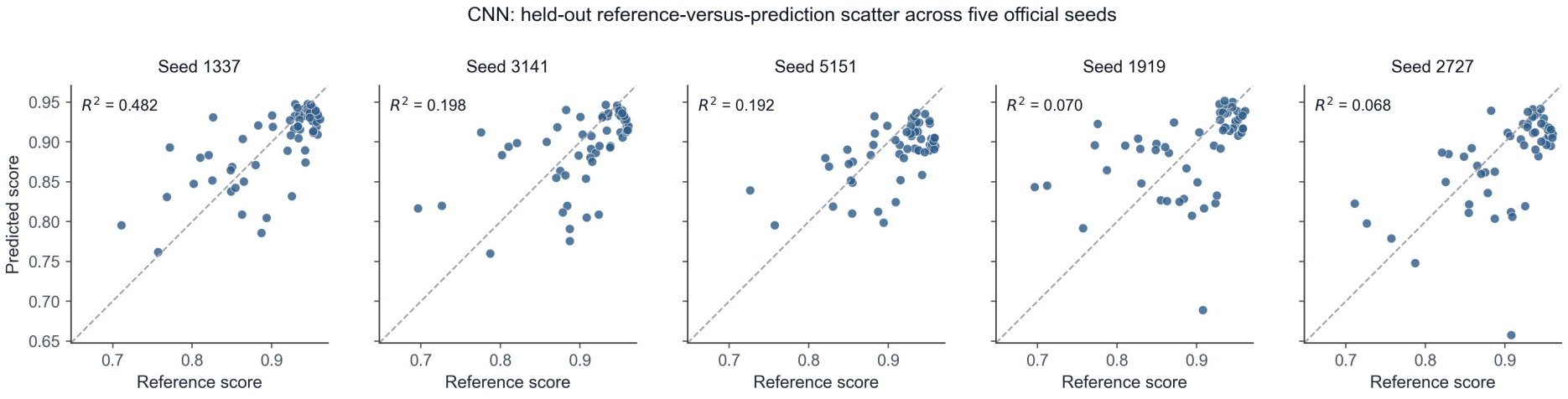

**Fig. S12 CNN held-out reference-versus-prediction scatter across the five official seeds, with per-panel $R^2$ annotations.**

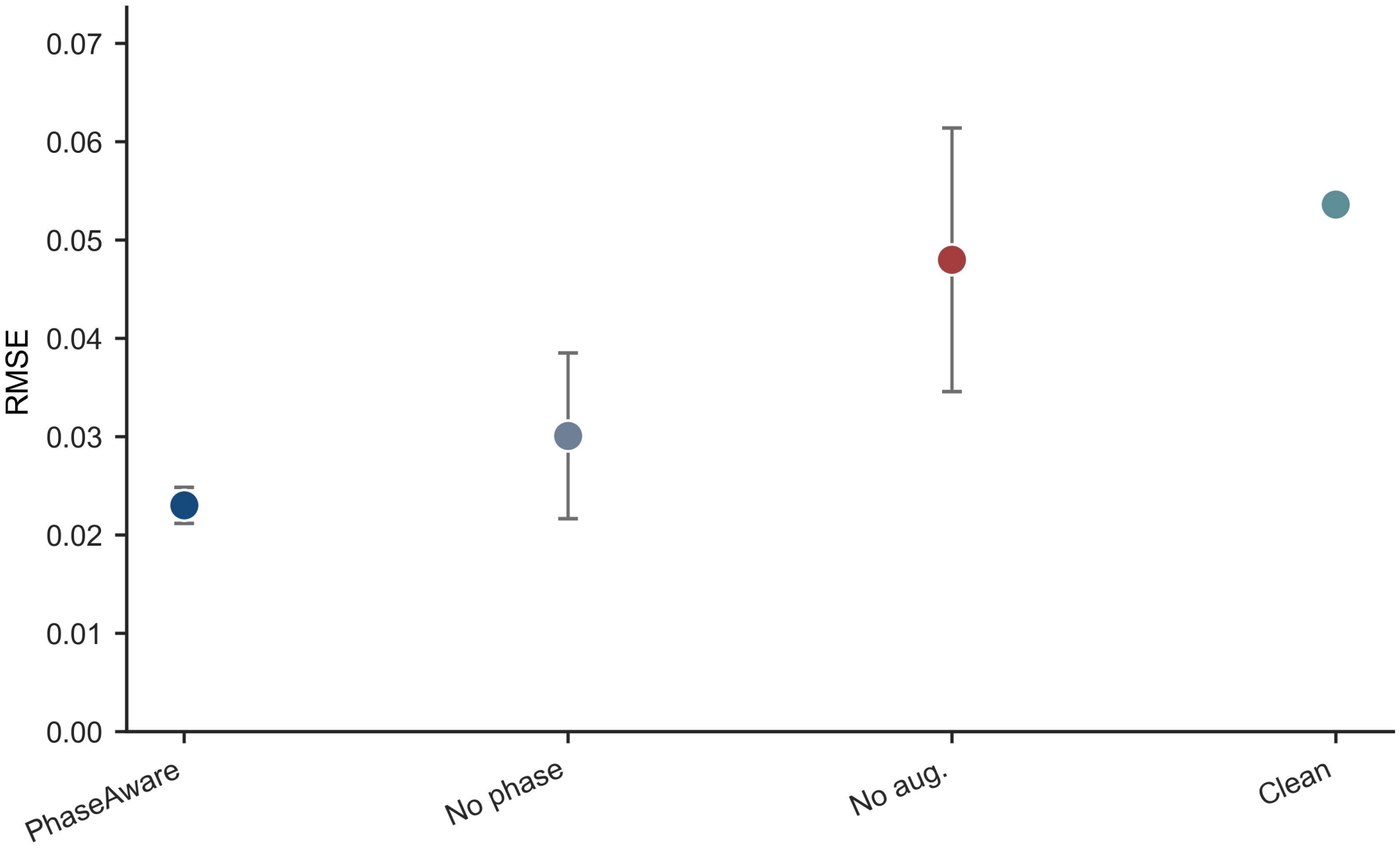

**Fig. S13 Supplementary ablation sensitivity for RMSE. Points show mean ± SD without connecting lines.**

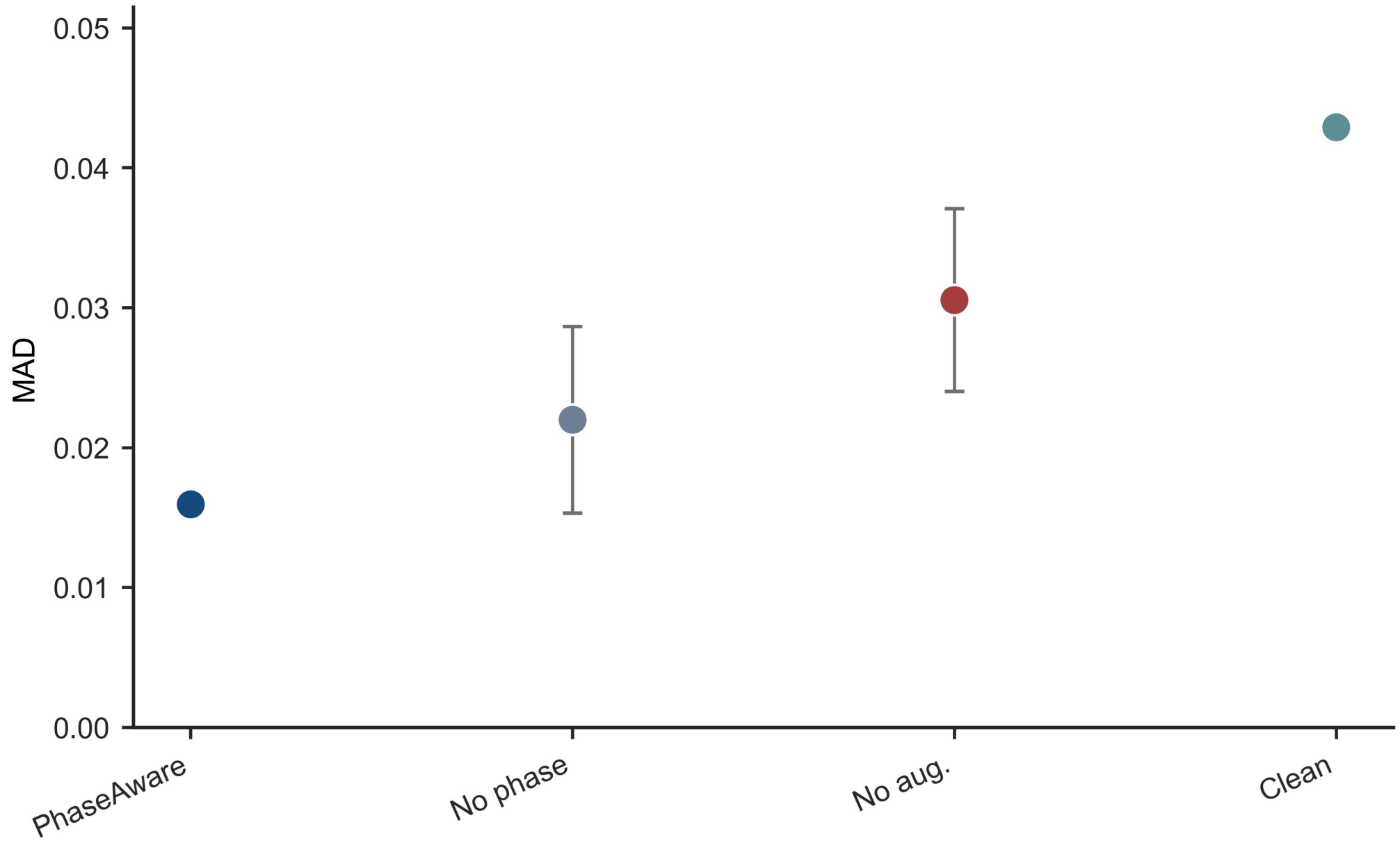

**Fig. S14 Supplementary ablation sensitivity for MAD. Points show mean ± SD without connecting lines.**

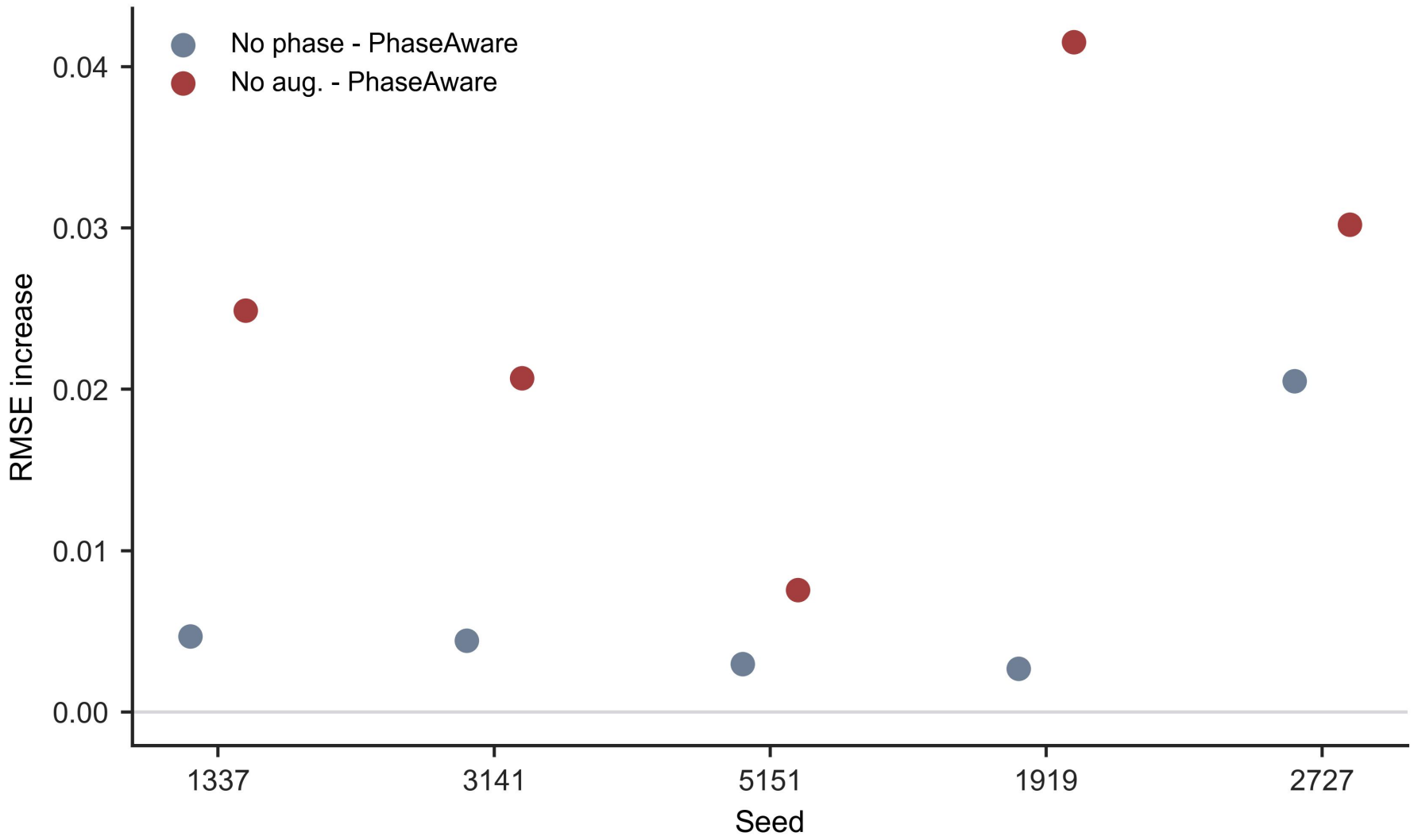

**Fig. S15 Seed-level mechanism deltas. Scatter points report RMSE increases relative to PhaseAware under matched seeds.**

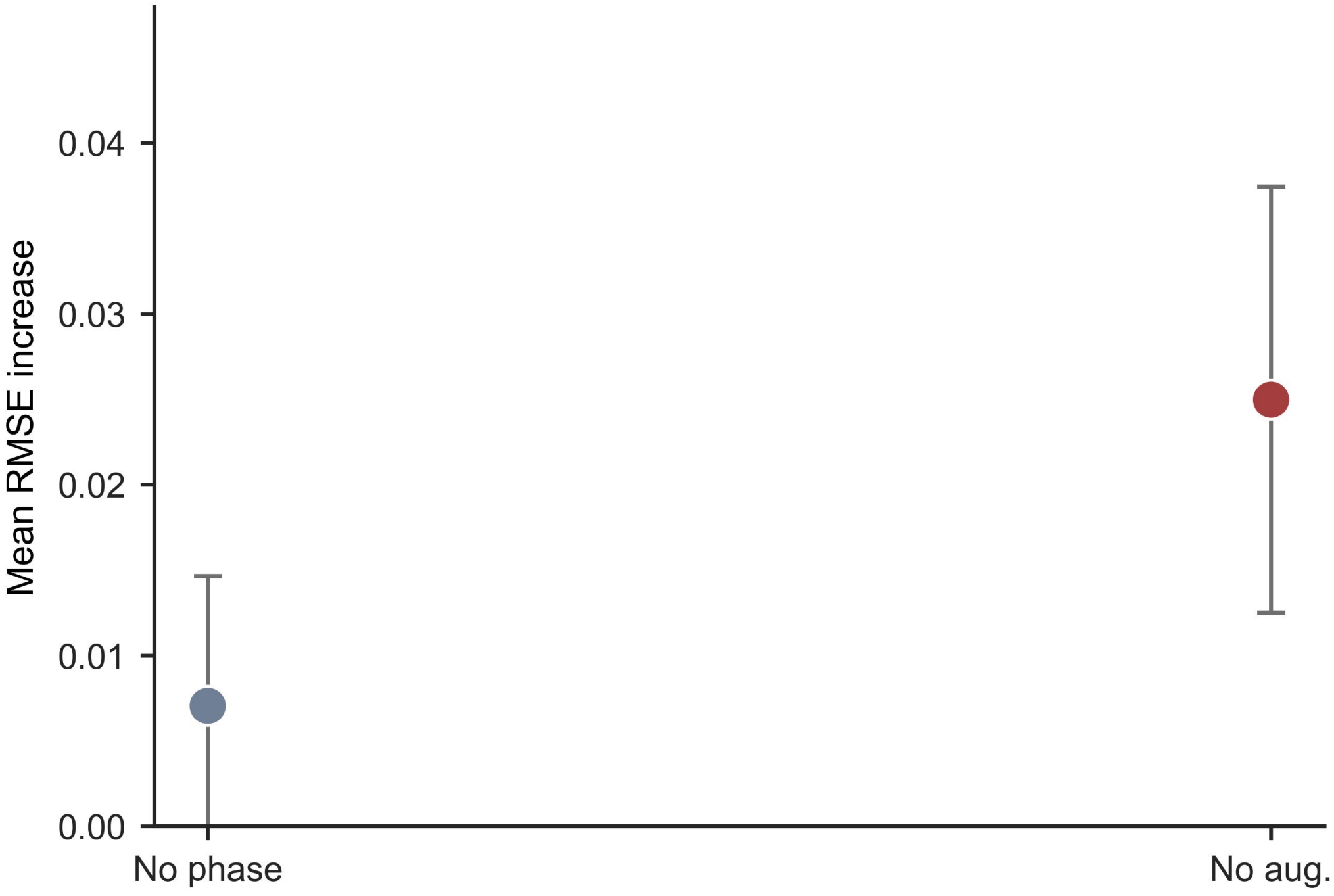

**Fig. S16 Average mechanism penalty. Points report mean RMSE increase and error bars report seed-level SD.**

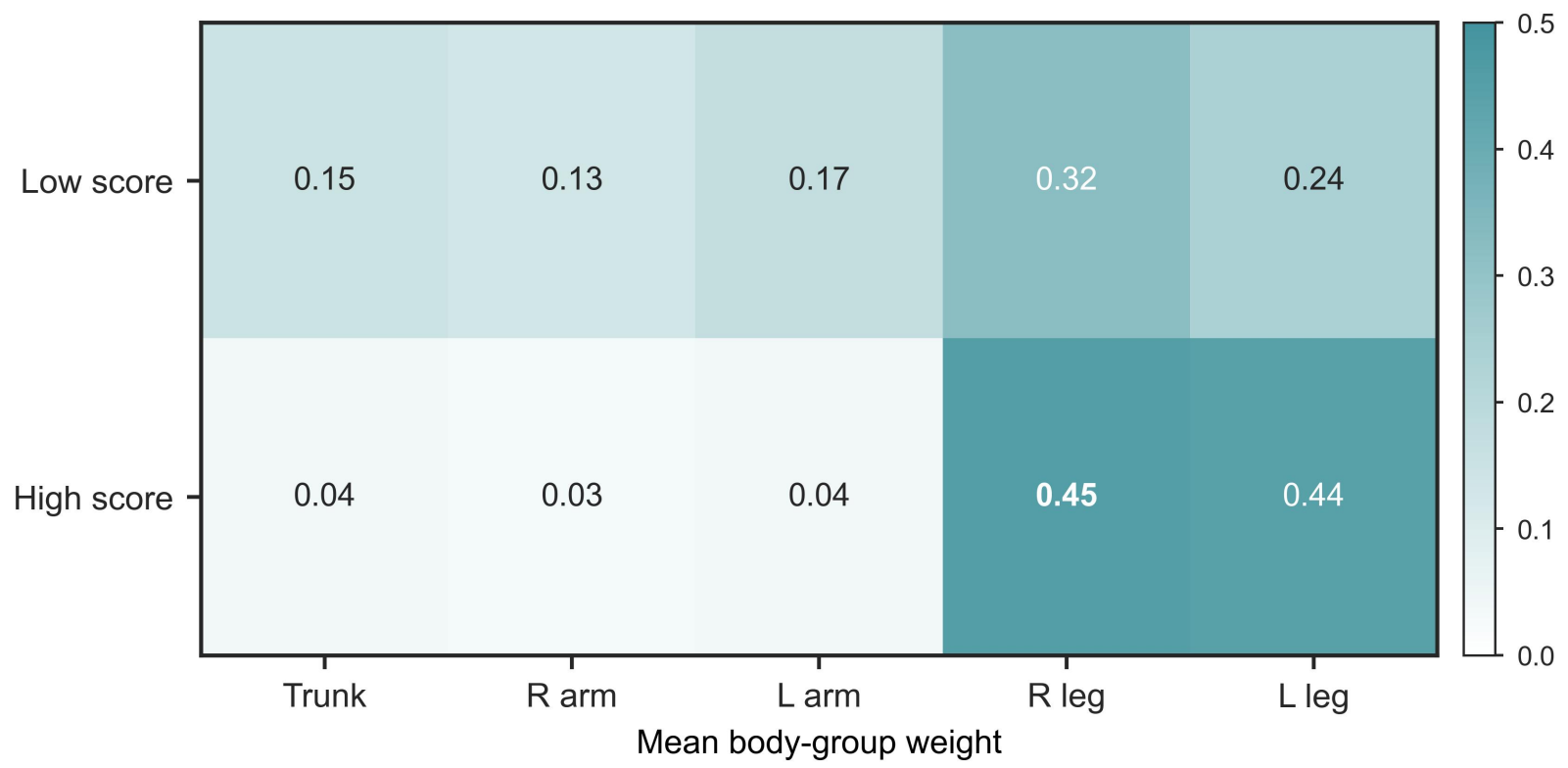

**Fig. S17 Supplementary body-weight heatmap. Rows compare low- and high-score bins.**

## Supplementary Note

This supplementary information is intended for inspection and reproducibility support. The main manuscript remains the authority for the final interpretation, clinical boundary, and workflow-facing claim.